\documentclass[10pt,twocolumn,letterpaper]{article}
\pdfoutput=1
\usepackage{wacv}
\usepackage{times}
\usepackage{epsfig}
\usepackage{graphicx}
\usepackage{epstopdf}
\usepackage{amsmath}
\usepackage{amssymb, multirow}

\DeclareGraphicsExtensions{.eps}
\graphicspath{ {images/} }

\newcommand{\squeezeup}{\vspace{-2.5mm}}

\wacvfinalcopy 

\def\wacvPaperID{275} 

\ifwacvfinal\fi
\begin{document}

\title{Detecting Temporally Consistent Objects in Videos through Object Class Label Propagation}
%
\author{Subarna Tripathi\\
UCSD\\
{\tt\small stripathi@ucsd.edu}
\and
Serge Belongie\\
Cornell University\\
{\tt\small sjb344@cornell.edu}
\and
Youngbae Hwang\\
KETI\\
{\tt\small ybhwang@keti.re.kr}
\and
Truong Nguyen\\
UCSD\\
{\tt\small tqn001@eng.ucsd.edu}
}

\maketitle

\begin{abstract}
Object proposals for detecting moving or static video objects need to address issues such as speed, memory complexity and temporal consistency. 
We propose an efficient Video Object Proposal (VOP) generation method and show its efficacy in learning a better video object detector. A deep-learning based  
video object detector learned using the proposed VOP achieves state-of-the-art detection performance on the Youtube-Objects dataset. 
We further propose a clustering of VOPs which can efficiently be used for detecting objects in video in a streaming fashion. As opposed to applying per-frame convolutional neural network (CNN) based object detection, our proposed method called Objects in Video Enabler thRough LAbel Propagation (OVERLAP) needs to classify only a small fraction of all candidate proposals in every video frame through streaming clustering of object proposals and class-label propagation. Source code will be made available upon publication.
     
\end{abstract}
\squeezeup

\section{Introduction}

Object proposal generation is a common pre-processing step for object detection in images, which is a key challenge in computer vision. Object proposals dramatically decrease the number of detection hypotheses to be assessed. Thus, use of CNN-features \cite{girshick14CVPR}, which is more effective but computationally expensive, have turned out to be feasible for accurate detection. 
For the detection of video objects, proposals not only need to consider the space-time complexity, but also need to address the temporal consistency. 
We propose generating Video Object Proposals (VOP) by scoring candidate windows based on spatio-temporal edge content and show that these VOPs help in learning  better video object detectors. Further, we propose an efficient online clustering of these proposals in order to process arbitrary long videos. 
\begin{figure}
\begin{center}
	\includegraphics[scale=0.45]{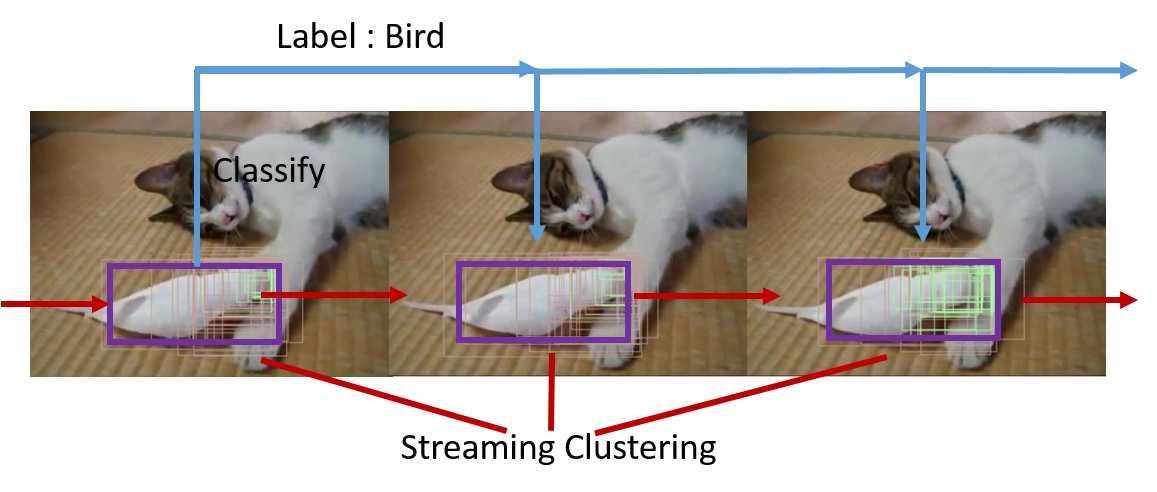}
\end{center}
\setlength{\abovecaptionskip}{12pt} 
   \caption{OVERLAP: Object class labels get propagated by streaming clustering of temporally consistent proposals. The system classifies only those VOPs which belong to a new cluster.}
\label{OVERLAP framework}
\squeezeup 
\end{figure} 
We show that the joint-analysis of all such windows provides a way towards multiple object segmentations and helps in reducing test time object detection complexity. 

We divide a video into sub-sequences with one-frame overlap in a streaming fashion \cite{XuXiCoECCV2012, StreamGBHppST14}. We analyze all candidate windows jointly within a sub-sequence followed by affinity-based clustering to produce temporally consistent clusters of object proposals at every video frame. 
The advantage of performing a streaming spatio-temporal clustering on the object proposals is that it enables an easy label propagation through the video in an online framework. Presumably all object proposals of a cluster have the same object class type. We propose deep-learning based video object detection through objects' class label propagation using online clustering of VOPs. 
As opposed to applying R-CNN \cite{girshick14CVPR}-like approaches, which essentially classify every window based on the expensive CNN features, at every video frame, the proposed method of label propagation requires detection/classification only on video frames which has new clusters
(fig. \ref{OVERLAP framework})
.
Our main contributions are as follows:
\squeezeup 
\begin{itemize}
  \item We present a simple yet effective Video Object Proposal (VOP) method for detecting moving and static video objects by quantifying the spatio-temporal edge contents;
  \squeezeup
  \item We present a \textbf{novel algorithm}, ``Objects in Video Enabler thRough LAbel Propagation'' (OVERLAP), that exploits objects' class label propagation through streaming clustering of VOPs to efficiently detect objects in video with  temporal consistency. 
\end{itemize}
\squeezeup
We also present the following minor contributions:
\squeezeup 
\begin{itemize}
  \item Demonstrating VOP's efficacy in learning a better CNN-based video object detector model;
  \squeezeup 
  \item Object segmentation as a by-product of object detection framework, OVERLAP. 
\end{itemize}




\section{Related Works}

There are several approaches towards video object detection which broadly fall into three categories : (1) image object proposals for each frame (2) motion segmentation in video (3) supervoxel aggregation. 

\textbf{Object Proposals and detection:} 
Unsupervised category-independent detection proposals are evidently shown to be effective for object detection in images. 
Some of these methods are Objectness \cite{Objectness2010CVPR}, category-independent object proposals \cite{Endres10ECCV}, SelectiveSearch \cite{SelectiveSearch2013IJCV}, MCG \cite{APBMM2014}, GOP \cite{GOP2014}, BING \cite{BingObj2014}, EdgeBoxes  \cite{Dollar2014ECCV}. A comparative literature survey on object proposal methods and their evaluations can be found in \cite{Hosang2014Bmvc,Hosang2015pami}. 
Although there is no ``best'' detection proposal method, EdgeBoxes, which scores windows based on edge content, achieve better balance between recall and repeatabilty. 

Applying image object proposals directly for each frame in video may be problematic due to time complexity and temporal consistency. In addition, issues like motion blur and compression artifacts can pose significant obstacles to identifying spatial contours, which degrades the object proposal qualities.
Recent advances like SPPnet \cite{SPP_NET_HeZR014}, Fast R-CNN \cite{fast_RCNN_15}, and Faster R-CNN \cite{Faster_RCNN_RenHG015} have dramatically reduced the running time by computing deep features for all image locations at the same time and snapping them on appropriate proposal boxes. Per-frame object detection still needs classification of proposal windows and temporal consistency still remains a challenge. The proposed framework dispenses with the need of classifying every candidate window of every video frame through  spatio-temporal clustering, thus addressing temporal consistency.

\textbf{Motion Segmentation in Video:}
Motion based segmentation is the task of separating moving foreground objects from the background. Several popular methods of motion segmentation include the  layered Directed Acyclic Graph (DAG) based framework \cite{Zhang13CVPR}, Maximal weight cliques \cite{Ma12CVPR}, fast motion segmentation \cite{Papazoglou13ICCV}, tracking many segments \cite{Li_2013_ICCV}, identifying key segments \cite{key_segments_Lee11}, and many more. 
Although video motion segmentation can detect moving foregrounds robustly, it is not easy to detect multiple objects or if objects suddenly stop or change motion abruptly.

\textbf{Supervoxel aggregation:}
Spatio-temporal object proposals have been considered in the context of aggregating supervoxels with spatio-temporal connectivity between neighboring labels. Jain {\em et al} \cite{Jain14CVPR} developed an extension of the hierarchical clustering method of SelectiveSearch \cite{SelectiveSearch2013IJCV} to obtain object proposals in video. Even though their independent motion evidence effectively segment objects with motions from the background, static objects can not be recovered. Oneata {\em et al.} \cite{SpTmp-Obj_prop14} presented spatio-temporal object proposals by a randomized supervoxel merging process. 
Sharir {\em et al.} \cite{VidObjProp12} proposed the extension of category-independent object proposals \cite{Endres10ECCV} from image to video by extracting object proposals at each frame and linking across frames into object hypotheses in a framework of graph-based segmentation using higher-order potentials leading to a high computational expense.
All these supervoxels often cannot replace the object proposal step for object detection either due to its complexity or the associated over-segmentations. 






\section{Video Object Proposals}

We extend EdgeBoxes \cite{Dollar2014ECCV} from generating image object proposals to video object proposals. 
In addition to the spatial edge responses, $\mathbf{E_s}$, at every pixel in EdgeBoxes \cite{Dollar2014ECCV}, we consider exploiting temporal edge responses, $\mathbf{E_t}$, at every pixel location using mid-range optical flow analysis. 
\squeezeup 
\begin{equation} \label{edge_define}
\mathbf{E_t}, \mathbf{E_t} \in \mathbb{R}^{M\times N}_{\geq 0}
\end{equation}
$M$ and $N$ are the height and width of an image. 

\subsection{Spatio-Temporal Contours and VOP} \label{spatio-temp-contours}

Optical flow field between pairs of consecutive frames provide approximate  motion contours. We perform 2-frame forward optical flow \cite{TBroxOF11} for every consecutive frame-pair. At every pixel, the magnitude of the flow field's gradient and the difference in direction of motion from its neighbor \cite{Papazoglou13ICCV}, contribute to the measure of the motion contour. To address incompleteness and inaccuracies of two-frame optical flow estimation, we analyze mid-range optical flow over a subset of video frames. Within a sub-sequence, we approximate which pixels consistently reside inside a moving  object using inside-outside maps \cite{Papazoglou13ICCV}. 
In our experiments, the inside-outside maps, accumulated over 3 - 5 frames, provide good estimates of time-consistent gross location priors for moving objects. In Section \ref{Streaming_Clustering}, We describe how this mid-range accumulation can effectively be exploited for object label propagation through streaming clustering. A simple edge detector on this location prior is called temporal edge, $\mathbf{E_t}$.   

\begin{figure}
\begin{center}
	\includegraphics[scale=0.42]{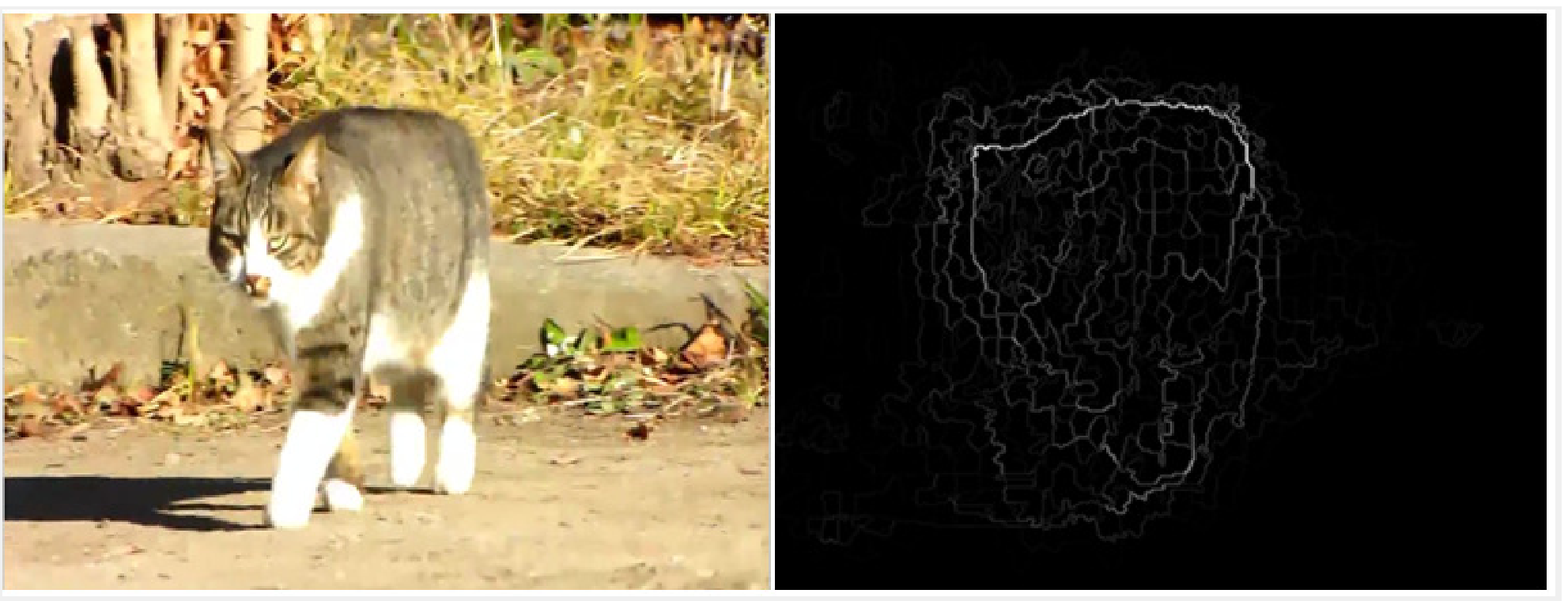}
	\includegraphics[scale=0.42]{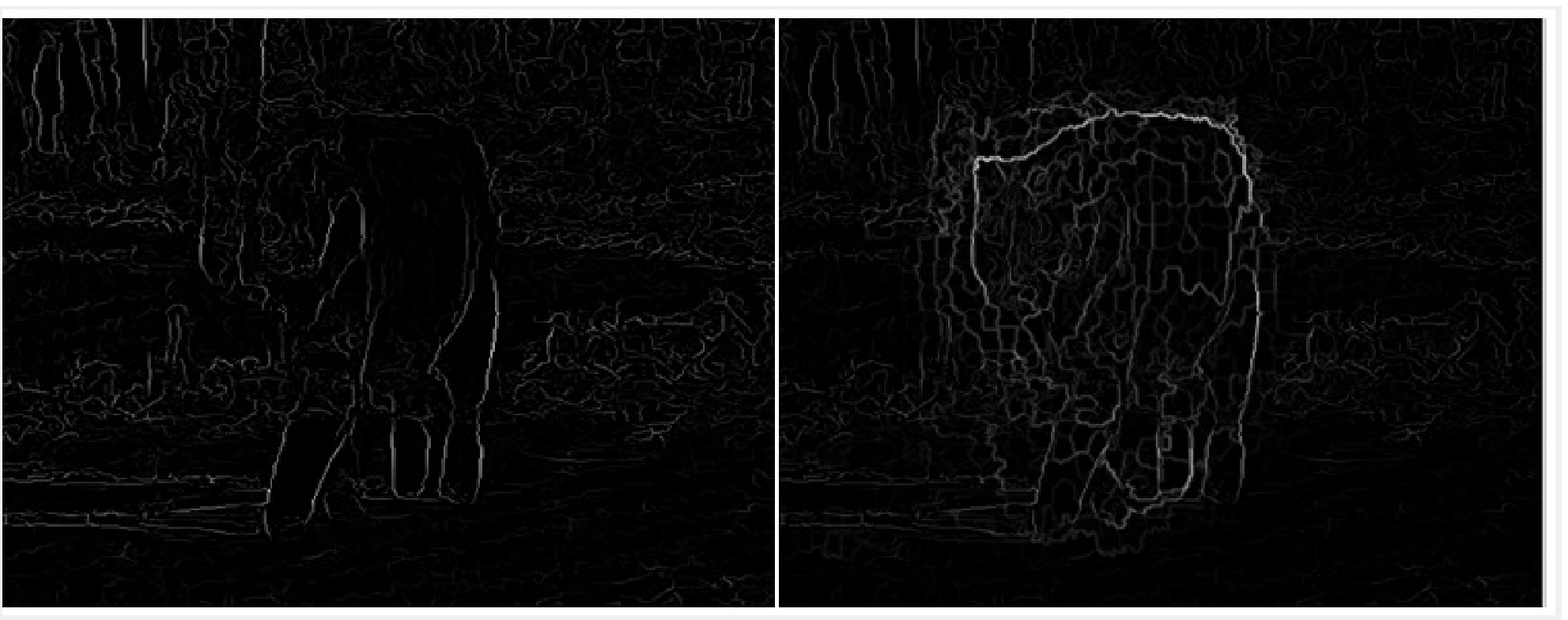}
\end{center}
   \caption{Spatio-Temporal EdgeBoxes. Clock-wise from top-left - (i) A video frame from Youtube-Objects dataset, (ii) Temporal edge $\mathbf{E}_t$ \ie normalized gradient of location prior from motion analysis, (iii) Spatial edge $\mathbf{E}_s$ \ie structured edge detection on video frame, (iv) linear combination of spatial and temporal edge $\mathbf{E}$ brings out the prominent contours of both static and/or moving objects.  }
\label{fig:spatio-temporal-edgebox}
\end{figure}

EdgeBoxes \cite{Dollar2014ECCV} employs efficient data structures to score millions of candidates based on the difference of number of spatial contours  that exist in the box and those that straddle box's boundary. We use a similar scoring strategy, but on spatio-temporal edge $\mathbf{E} \in \mathbb{R}^{M\times N}_{\geq 0}$ which is formed according to Eq \ref{spatio-temp-edge}. 
\squeezeup 
\begin{equation} \label{spatio-temp-edge}
\mathbf{E = \lambda E_t + (1-\lambda )E_s} , \mathbf{\lambda} \in [0,1]
\end{equation}
As the value of $\lambda$ increases, the system favors detecting only moving objects. One example of spatio-temporal edge is demonstrated in figure \ref{fig:spatio-temporal-edgebox}. A linear combination of spatial and temporal edge responses represents spatio-temporal contours. This enables a simple yet efficient strategy for scoring based on spatio-temporal edge content through edge groups. We find intersecting edge groups along horizontal and vertical boundaries using two efficient data structures\cite{Dollar2014ECCV}. We use the integral image-based implementation to speed up scoring of boxes in sliding window fashion.
As described in the next section, object proposals based on these spatial-temporal contours outperforms those based on only spatial contours for video object detection. The presence of motion blur in spatial edges affects the performance of spatial contour based proposal prediction in video frames. In practice, $\lambda=0.2$ to $0.5$ works well for Youtube-Objects dataset. 


\section{Learning Video Object Detector Model}

We aim for detecting objects in generic consumer videos. Due to the domain shift issues between images and video frames \cite{KalogeitonFS15}, our 10-class video object detection uses supervised pre-training from ImageNet reference model for classification and fine-tuning on annotated frames from Youtube-Objects dataset v2.0 \cite{youtube-Objects, KalogeitonFS15, ObjClsDet2012} for video objects detection.  

\textbf{Youtube-Objects dataset.} The dataset is composed of videos collected from Youtube by querying for the names of 10 object classes of the PASCAL VOC Challenge. It contains 155 videos in total and between 9 and 24 videos for each class. The duration of each video varies between 30 seconds and 3 minutes. However, only $6087$ frames are annotated with a bounding-box around an object instance. Hence, the number of annotated samples is approximately 4 times smaller than in PASCAL VOC.

The bottom-up region proposal methods play an important role. Motion blur and compression artifacts affects the quality of spatial edges in video frames, thus, generating good object proposals becomes more challenging. This is to be noted that R-CNN \cite{girshick14CVPR}, or, Fast R-CNN \cite{fast_RCNN_15} are fine-tuned for image object detection task, especially for 20-class PASCAL VOC image object categories which is a superset of Youtube-objects categories\textit{}.

\textbf{Feature extraction.} We extract a 4096-dimensional feature vector corresponding to each region proposal using GPU-based (GeForce GTX 680) the Caffe \cite{Caffe13} implementation of the CNN described by Krizhevsky \etal \cite{AlexNet12}. Features are computed by forward propagating a mean-subtracted 227 $\times$ 227 R-G-B image through five convolutional layers and two fully connected layers.

\textbf{Region Proposals.} We use approximately $2000$ candidate proposals per video frame to be processed for learning detectors. We investigate the object detection model with different region proposal methods such as selective search\cite{SelectiveSearch2013IJCV}, EdgeBoxes\cite{Dollar2014ECCV}. As the resolution of different videos varies from VGA to HD, we re-size every video frame to $500\times 500$ before performing proposal generation task. 

\textbf{Training.} 
We discriminatively pre-train the CNN on a large auxiliary dataset (ILSVRC2012 classification) using image-level annotations, followed by domain specific fine-tuning by replacing the last layer of AlexNet \cite{AlexNet12} model with $10+1$ softmax output layer. We use two-step initialization for fine-tuning as described in \cite{InitializationTip14}. As per PASCAL detection criteria, we treat all region proposals with $\geq$ 0.5 IoU overlap with a ground-truth box as positives for that class of the box and the rest as negatives. Once features are extracted and training labels are applied, we optimize one linear SVM per class. 

\textbf{Test time detection.} During test time, approximately $500$ to $2000$ VOPs are generated. Then forward propagation is performed through the CNN to compute features. Finally, we perform scoring using per-class trained SVM similar to \cite{girshick14CVPR} followed by non-maximum suppression.  

In the below section \ref{OVERLAP}, we describe the alternate test time detection using OVERLAP through objects' class label propagation. 



\section{OVERLAP: Objects in Video Enabler thRough LAbel Propagation } \label{OVERLAP}

Classical approach for object localization has traditionally been image window classification, where each window is scored independent of other candidate windows. Recently, more success in object detection has been reported by considering spatial relations to all other windows and their appearance similarity \cite{VezhnevetsF15} with examplar-based associative embedding. 
Our approach towards video object detection considers spatial relationship and appearance similarity with windows within and even in other nearby video frames, yet in much simpler way through spatio-temporal clustering, to avoid classifying every candidate windows. 
\subsection{Joint analysis of windows}

We aim to detect dissimilarity between VOPs generated within a sub-sequence based on a simple underlying principle: proposals corresponding to the same object exhibit higher statistical dependencies than proposals belonging to different objects. 

As a motivation for our approach, we consider generating proposal boxes by perturbing the ground truth locations of PASCAL VOC 2007 objects' bounding boxes and observe the statistical association of those proposal boxes. Let, $A$ and $B$ denote generic features of neighboring proposal windows, where neighborhood is characterized by non-zero Intersection-Over-Union (u). We investigate the joint distribution over pairings ${A,B}$. Let, $\mathit{p}(A,B;u)$ be the joint probability of features A and B of windows with spatial overlap value, $u$. Then, $P(A,B)$ could ideally be computed as :
\squeezeup 
\begin{equation} \label{joint_density}
P(A,B) = \frac{1}{Z} \int_{u}^{1}  {w(u) \mathit{p}(A,B; u)du}
\end{equation}
where, $u \in [0,1]$, $w$ is a weighting function, $w(0) = 0$, and $Z$ is a normalization constant. To simplify the process, we use uniform weighting function and work in the discrete (quantized) space of $u$ and replace the integral with summation. We take the marginals of the distribution to get $P(A)$ and $P(B)$. Motivated by the  analysis presented in crisp boundary detection work by Isola \textit{\etal} \cite{crisp_boundaries}, we model affinity with point-wise mutual information like function: 
\squeezeup 
\begin{equation} \label{PMI}
PMI_{\rho}(A,B) = log \frac{P(A,B)^{\rho}} {P(A) P(B)}
\end{equation}

\begin{figure}
\begin{center}
\includegraphics[scale=0.55]{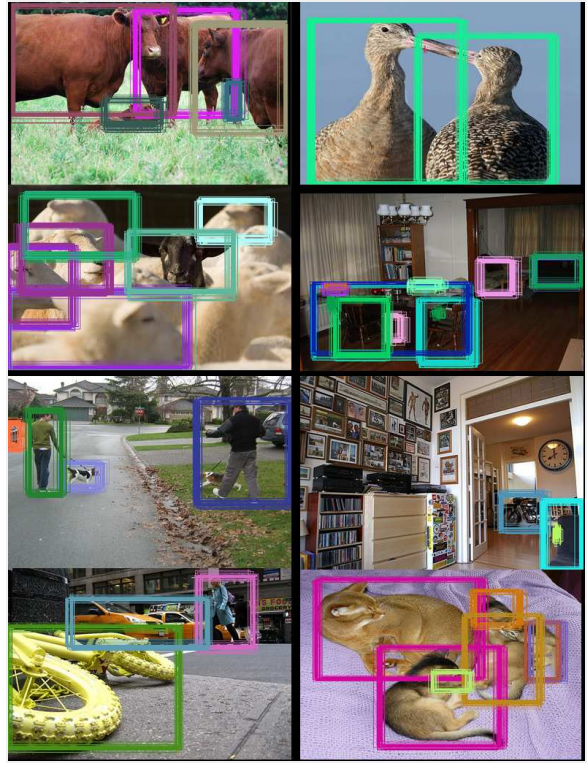}
\end{center}
   \caption{Synthetic experiment using perturbed ground truth locations in PASCAL VOC showing affinity-based clustering of candidate windows. The clustering technique works efficiently with some exceptions where overlapping object instances share very similar color. }
\label{fig:affinity-based-clustering}
\end{figure}

We choose the value of $\rho$ to be $1.2$ which produces best performance in PASCAL VOC dataset with perturbed ground truth proposals. In order to identify the boundary between two features, the model needs to be able to capture the low probability regions of P(A,B). We use a non-parametric kernel (Epanechnikov) density estimator. The number of sample points are the number of overlapping candidate windows. We perform affinity-based clustering afterwards. The affinity matrix, $\mathbf{W}$, for a sub-sequence is created from the affinity function, $PMI_{\rho}$, as follows:

\squeezeup 
\begin{equation} \label{affinity_matrix}
\mathbf{W}_{i,j} = e^{PMI_{\rho}(\mathbf{f}_i, \mathbf{f}_j)}
\end{equation}
Where $i$ and $j$ are the indices of proposal windows and $\mathbf{f}$ is the feature vector defined for a proposal window. 
Figure \ref{fig:affinity-based-clustering} shows an example of spatial clustering of proposal windows. The boxes drawn in same colors correspond to same clusters. 
In the streaming VOP clustering framework for Youtube videos, we perform the joint analysis on all proposals within a sub-sequence. Intuitively this is an easy yet effective clustering technique which works on actual test proposals (see Figure \ref{fig:Streaming-VOP}) which are not simply perturbed ground truth bounding boxes. 

This is to be noted that the proposed method measures the affinity between different object proposal windows (of varying sizes) within a video sub-sequence (in different video frames) unlike \cite{crisp_boundaries}, where the affinity is between the neighboring pixels in an image. 


\begin{figure}
\begin{center}
	\includegraphics[scale = 0.47]{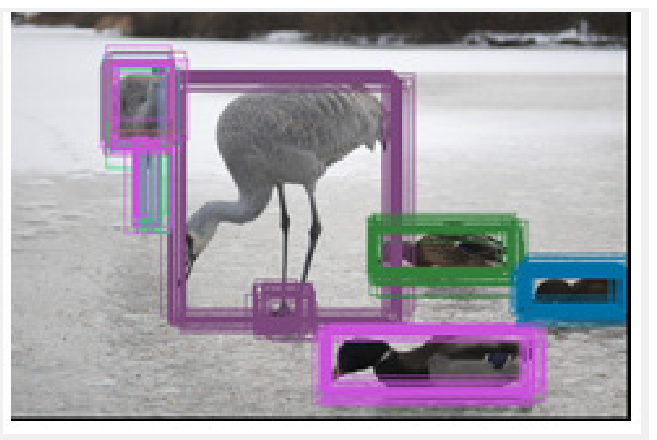}\\ \vspace{-0.1cm} 
	\includegraphics[scale = 0.42]{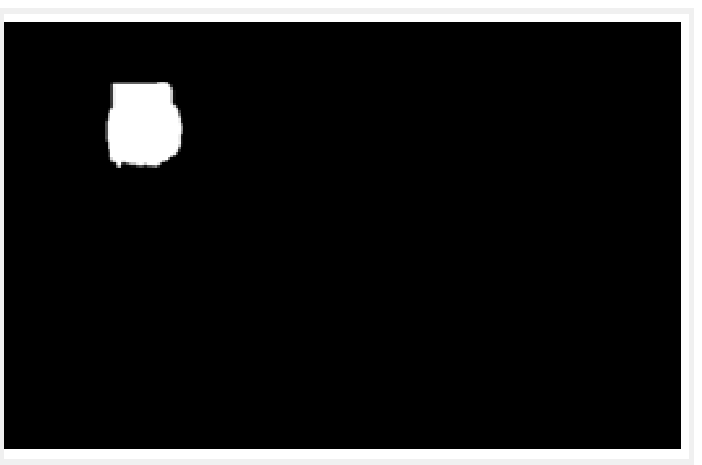} \includegraphics[scale = 0.42]{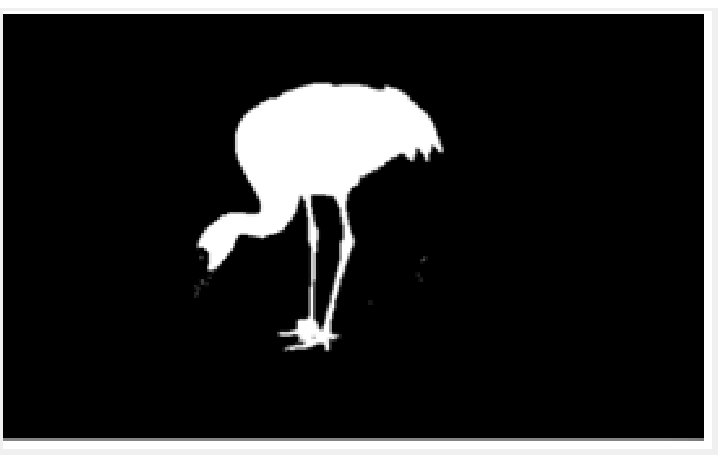}\\ \vspace{-0.1cm} 
	\includegraphics[scale = 0.42]{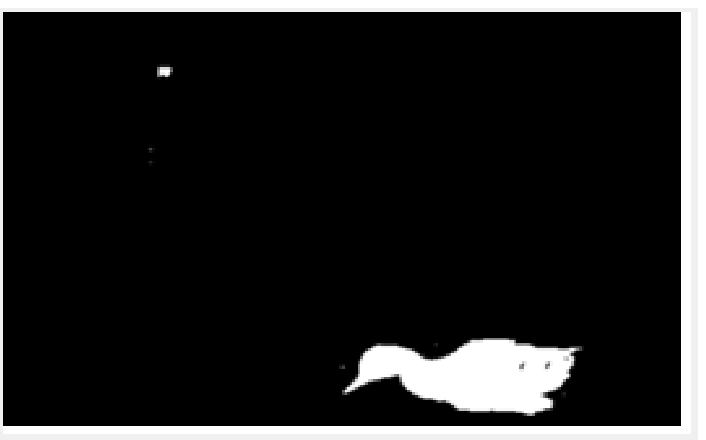} \includegraphics[scale = 0.42]{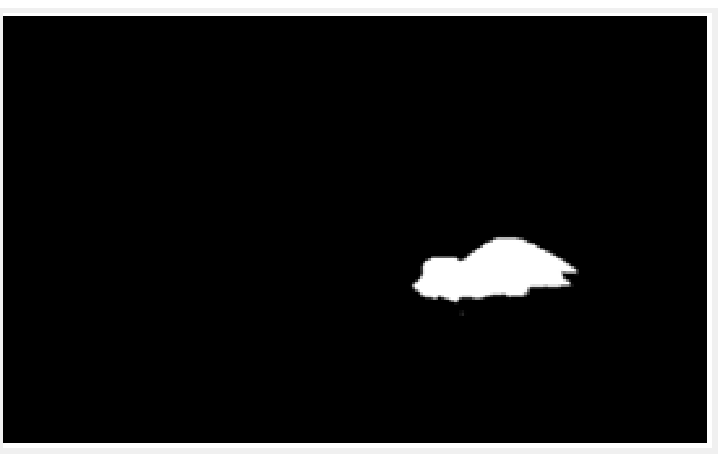}\\
	\vspace{0.1cm} 
	\includegraphics[scale = 0.47]{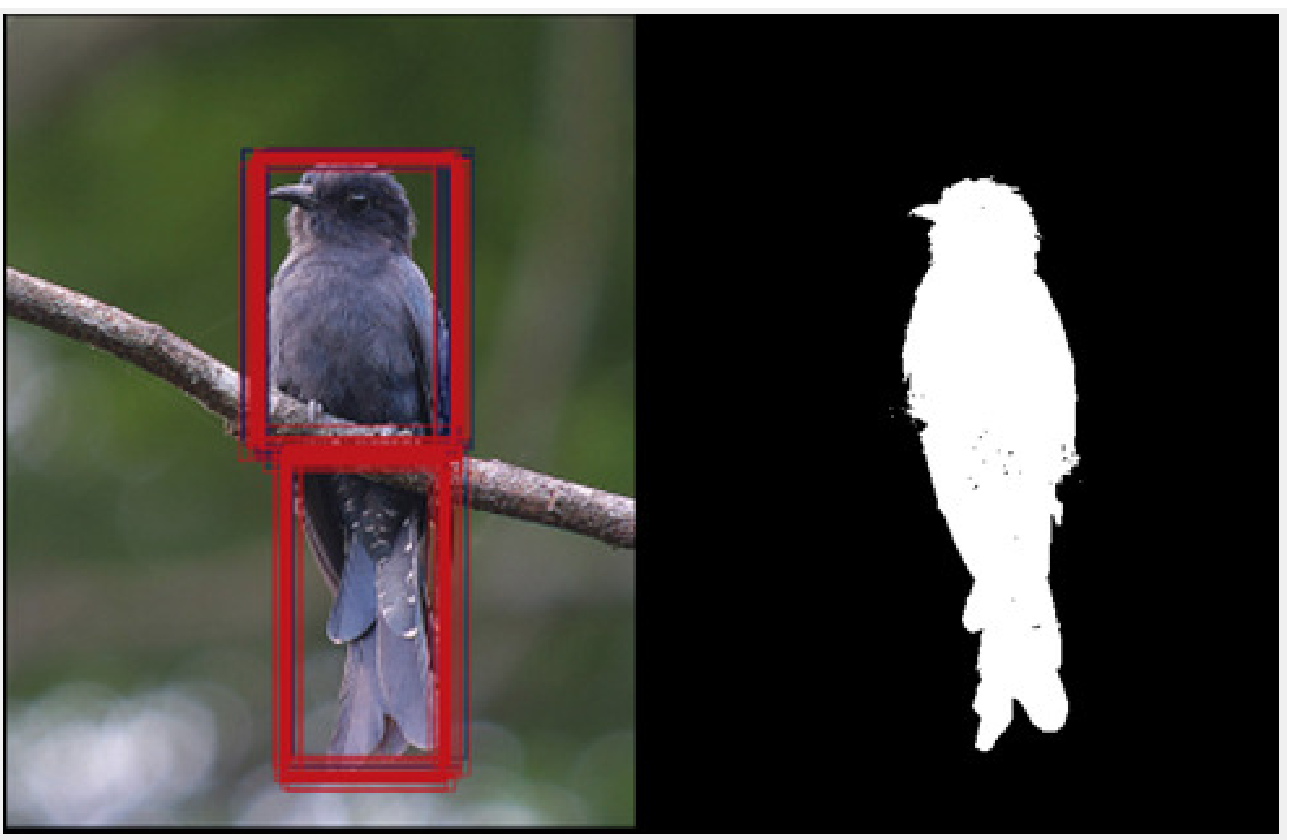}
\end{center}
   \caption{Segmentation masks from clustered object proposals. }
\label{fig:segmentation_mask1}
\squeezeup 
\end{figure} 

\begin{figure}
\begin{center}
	\includegraphics[scale = 0.34]{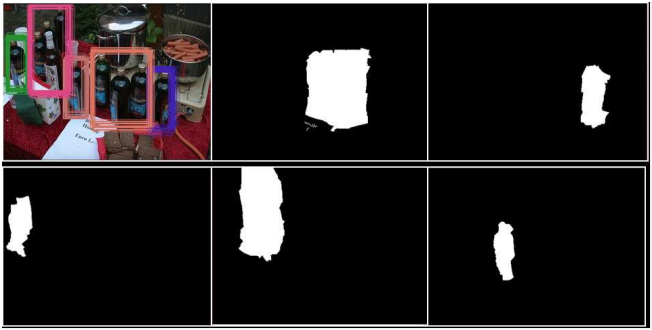}
\end{center}
   \caption{Segmentation masks generation is not successful where proposal boxes do not tightly enclose the actual object. This happens in cluttered background cases. }
\label{fig:segmentation_mask3}
\squeezeup 
\end{figure}

We demonstrate object segmentation can be achieved as a by-product of this clustering algorithm. We cast the segmentation problem through random-field based background-foreground segmentation without manual labeling \cite{SegProp12}.
Uniformly weighted sum of the location of every window corresponding to a unique cluster defines the foreground location prior for that cluster. 
However, unlike \cite{SegProp12}, in our approach, the location prior is not coming from the global neighbors of the image but from within itself and the  clustering allows multiple objects segmentations. The segmentation works well if the proposal boxes tightly enclose an object as shown in figure \ref{fig:segmentation_mask1} with some failure cases as shown in figure \ref{fig:segmentation_mask3} where the proposal boxes do not tightly enclose the object in a cluttered background.

\begin{figure}
\begin{center}
	\includegraphics[scale = 0.48]{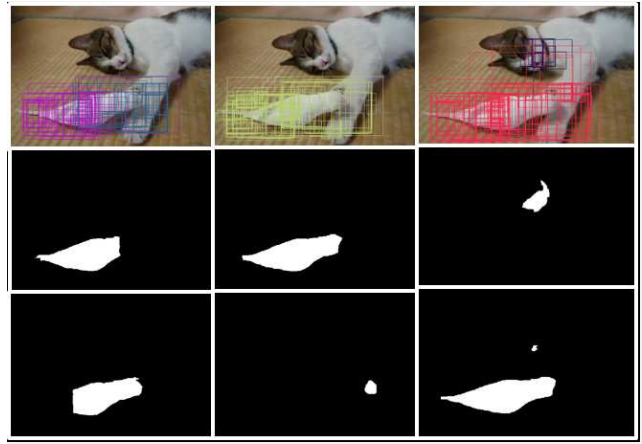}
\end{center}
   \caption{Two segmentation masks from clustered video object proposals from Youtube video ``Bird and Cat'' for frames \#5, \#20 and \#45. }
\label{fig:youtube_segmentation}
\squeezeup 
\end{figure}

Figure \ref{fig:youtube_segmentation} shows two segmentation masks generated on individual video frames from the clustered video object proposals generated by the proposed VOP on real videos.      

\subsection{Streaming clustering of proposals}  \label{Streaming_Clustering}
One of the main contributions of this paper is a simple, principled, and unsupervised approach to spatio-temporal grouping of candidate regions in streaming fashion. We describe a clustering framework which enforces a Markovian assumption on the video stream to approximate a batch grouping of  VOPs. A video is divided into a series of sub-sequences with one frame overlap with the previous sub-sequence as described in \cite{XuXiCoECCV2012}. VOP clustering within the current sub-sequence depends on the results from only the previous sub-sequence. We consider sub-sequence length of 3 to 5 frames as a trade-off between quality and complexity. This is the same sub-sequence volume, where mid-range motion analysis is performed for detecting temporal edges (Section \ref{spatio-temp-contours} ).
The color-histogram features are used for estimating joint probability between any overlapping window-pair and affinity-based clustering is performed afterwards. 
There are two important aspects in this streaming clustering method. The first is generic to any clustering algorithm \ie how to select the number of clusters and the second is specific to streaming method \ie how to associate cluster number of the current sub-sequence with any of the clusters of previous and/or future sub-sequences?
\squeezeup 

\subsubsection{Number of clusters.} Common consumer or Youtube videos contain limited number of moving objects, often less than five. Youtube-Objects dataset contains maximum of $3$ object instances and quite often a single moving object. We assume the presence of at most 5 objects to keep the computational complexity tractable and amenable to practical applications. We explore two modes of operations. The first method uses fixed number of clusters, $k=5$, with careful initialization of cluster centers using k-mean++ \cite{Kmeanspp07} during spectral clustering. The second one is the spectral clustering with self-tuning \cite{ZPClustering07}. We observe that while self-tuning outperforms the fixed cluster number case for hypothetically good object windows $($ such as the perturbed ground truth regions for PASCAL VOC $)$, both modes perform similarly in case of real object proposals generated by some proposal method. 
\squeezeup 
\subsubsection{Cluster Label Association.}
In the streaming framework, any sub-sequence except for the first one, needs to address the problem of either associating a cluster with a cluster number in the previous sub-sequence or generating a new one. We perform density estimation using an Epanechnikov kernel using the KD-tree implementation from \cite{KDE} for every cluster using the 4-dimensional location (2D center, height and width) and 45-bin color histogram (15 for each color channel) of the regions of the proposals corresponding to each and every cluster. If the minimum KL-divergence between a distribution of the current cluster and a cluster from the previous sub-sequence is less than a threshold, we perform the cluster assignment. Otherwise, we create a new cluster. 

This is to be noted that, for detecting primarily moving objects in videos, the weight of temporal edges could be as high as 0.7 or more as described in section \ref{spatio-temp-contours}. In such cases, considering as low as only 100 VOPs can potentially detect moving objects. Clusters may contain fewer number of proposal windows than the dimension of the original feature space which is 49-dimensional. Thus we perform PCA-based dimensionality reduction before estimating the distribution. Also, we perform scaling of features to ease the process of selecting kernel evaluation points.  
\subsection{Object Label Propagation.}

Time-consistent clustering enables object label propagation through the video. We perform CNN-based object detection \ie classification of every window in CNN feature space at a video frame only when we encounter a new cluster label. An assigned cluster label means the object category is the same as what was already detected in the associated cluster of the previous sub-sequence. We still need to perform the localization, however. In order to address the localization, we fit a 4-D Gaussian distribution on the location parameters \i.e. center (x,y), height(h) and width(w), of windows in a cluster. We simply keep track of the distance, $\mathbf{d}$, of the detected final object location (after first-time detection using R-CNN like approach) from the mean of the fitted Gaussian for every cluster. Furthermore, we localize the object by adding $\mathbf{d}$ with the mean of the 4-D Gaussian location distribution of the cluster in current sub-sequence. In general videos, new objects do not appear in every video frame. Thus, we do not need to detect objects at every video frame. Even when a new object appears, we need to detect/classify only for the proposals assigned to the new cluster. Thus, OVERLAP framework requires to process CNN features for only a small fraction of the number of proposals generated.

In some sense, the spatio-temporal clustering for object detection is related to tracking. However, a set of windows is tracked instead of a single region/object. Ability to bypass critical tracker initialization step and the possibility for applying R-CNN like detection at a more frequent and configurable interval to increase the detection accuracy (if needed) are the major advantages.

\section{Experimental Results}
\subsection{Video Object Detector using VOP}

We observe that the proposed VOP helps in learning a better object detector model. Table \ref{table:youtube-detection} shows the per-class detection accuracy and the mean Average Precision (mAP) for the 10-class Youtube-Objects Test set \cite{youtube-Objects}. 

\begin{figure*}[t]
\begin{center}
	\includegraphics[scale=0.8]{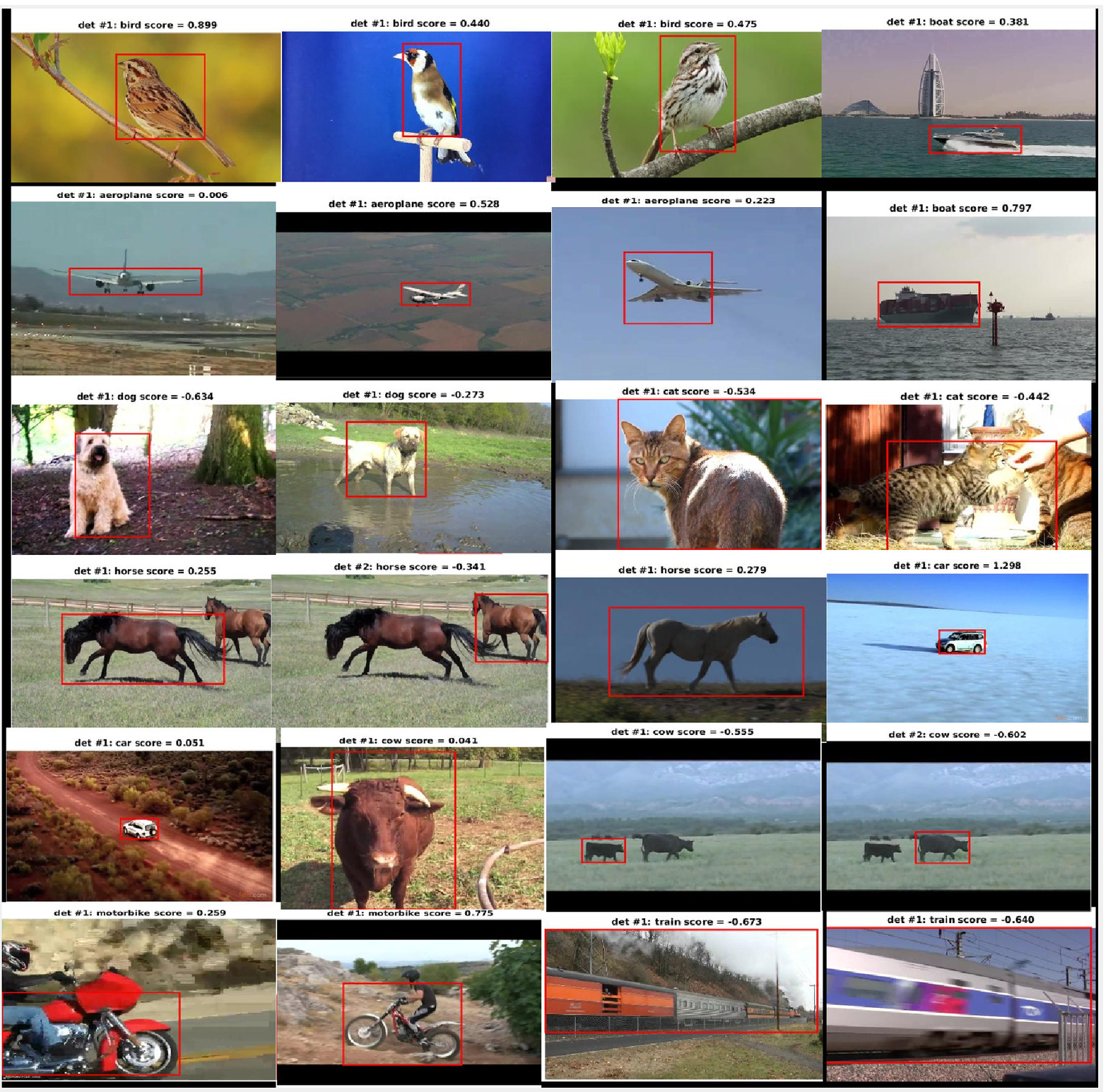}
\end{center}
\begin{center}
	\includegraphics[scale=0.3]{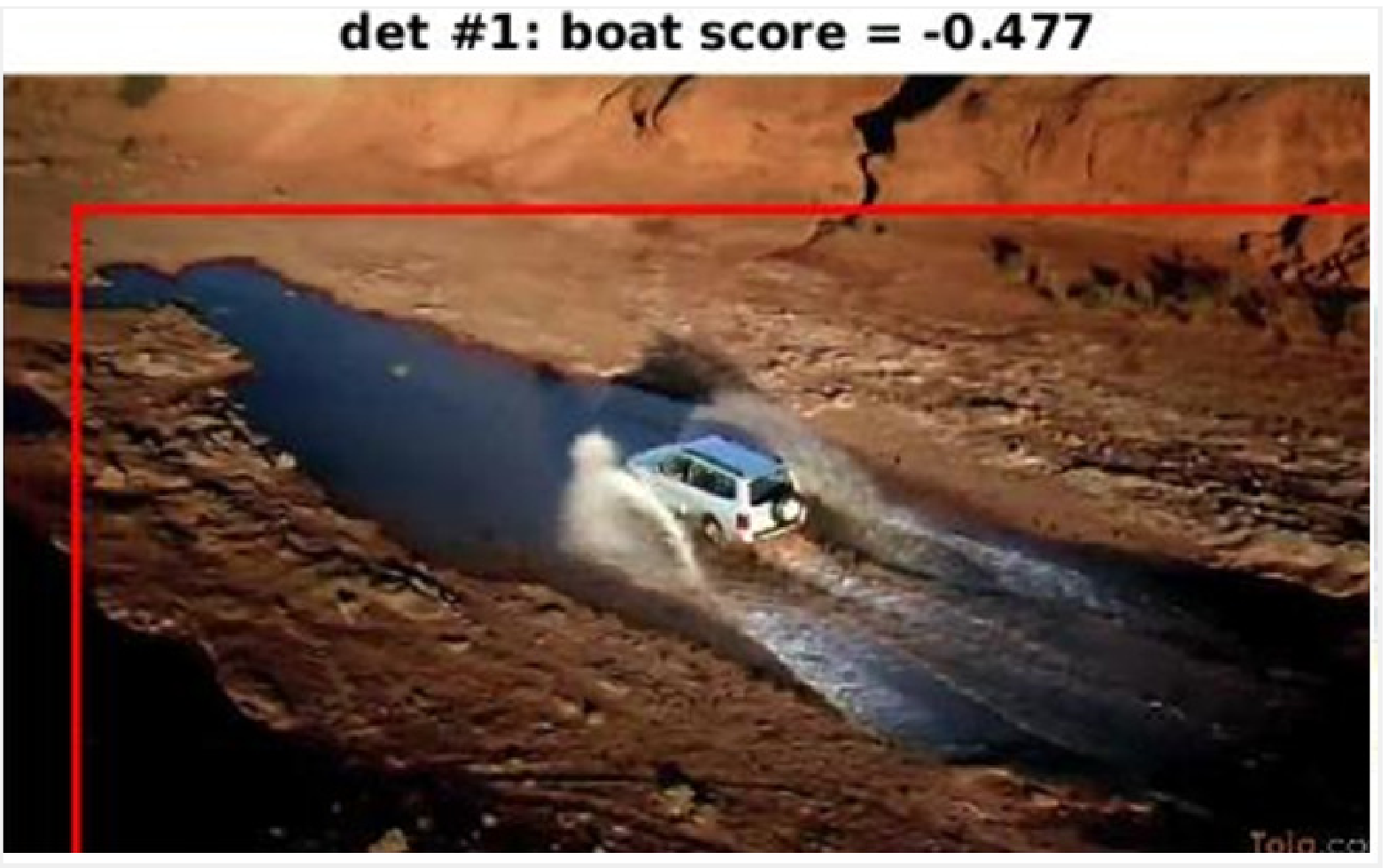}
	\includegraphics[scale=0.3]{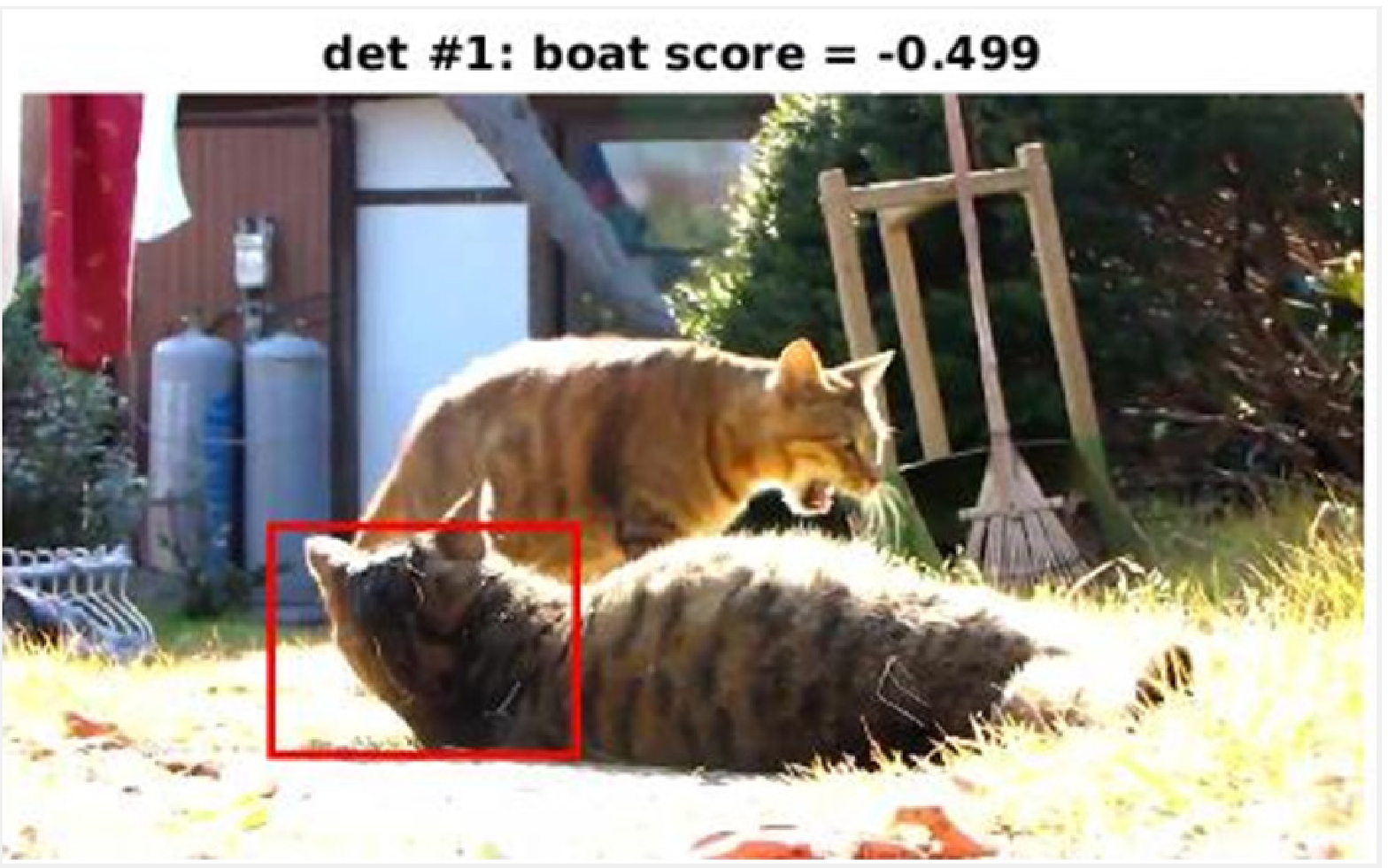}
	\includegraphics[scale=0.3]{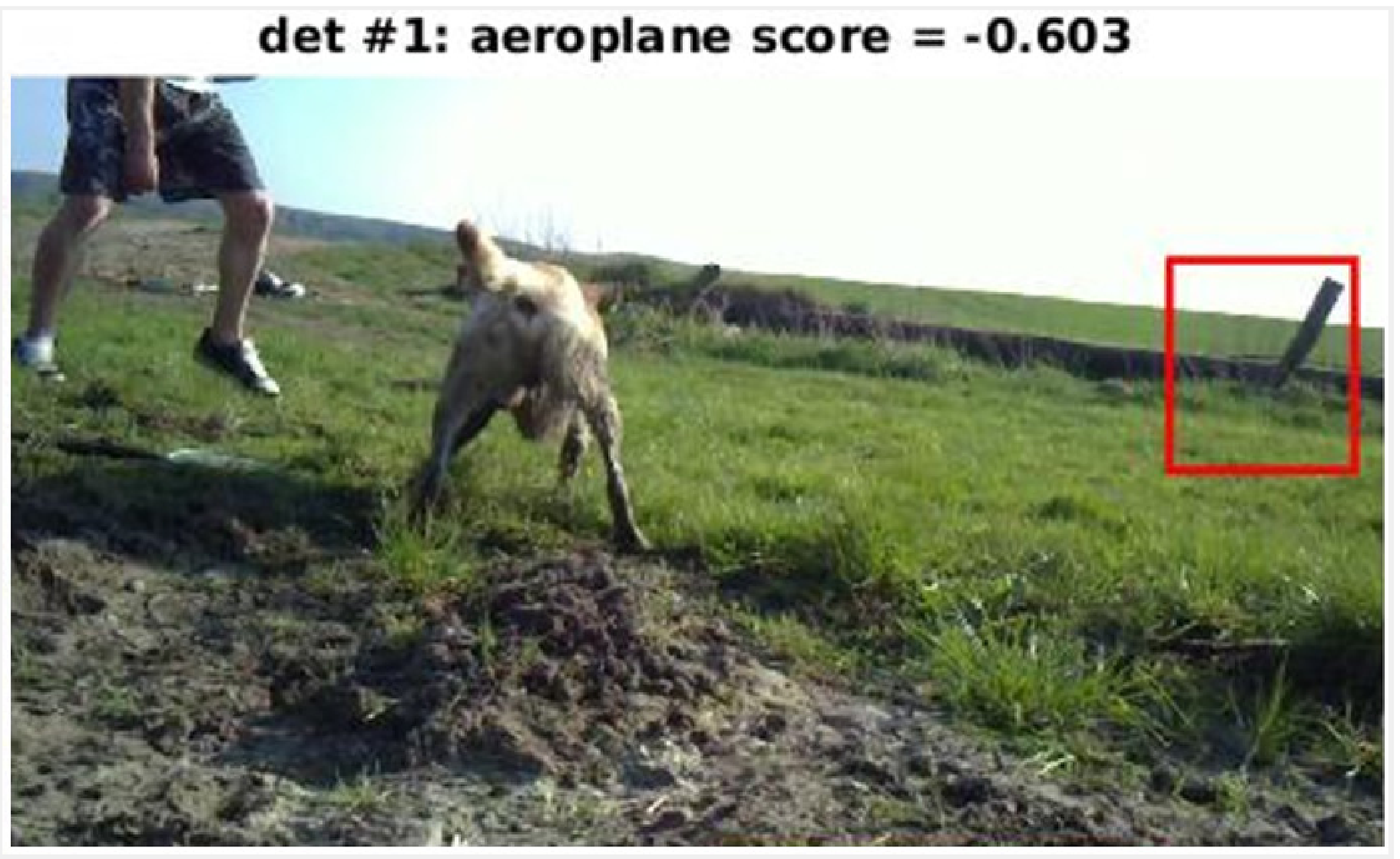}
\end{center}
   \caption{ Sample results of Video Object Detection with VOP. First 6 rows show successful detection cases and the last row shows false detection cases.}
\label{fig:Obj-det-VOP}
\squeezeup 
\end{figure*}

\begin{table}[!h]
\centering
\begin{tabular}{ |p{1cm}|p{0.8cm}|p{0.8cm}|p{1.1cm}|p{1.1cm}|p {1.1cm}| }
\hline
classes & R-CNN & DPM & Fine-tune SS & Fine-tune EB & Fine-tune VOP \\
\hline
plane & 14.1 & 28.42 & 25.57 & 26.52 & \textbf{29.77}\\ 
bird & 24.2 & \textbf{48.14} & 27.27 & 27.27 & 28.82\\ 
boat & 16.9 & 25.50 & 27.52 & 33.69 & \textbf{35.34}\\ 
car & 27.9 & \textbf{48.99} & 35.18 & 36 & 41\\ 
cat & 17.9 & 1.69 & 25.02 & 27.05 & \textbf{33.7}\\ 
cow & 28.6 & 19.24 & 43.01 & 44.76 & \textbf{57.56} \\ 
dog & 12.2 & 15.84 & 24.05 & 27.07 & \textbf{34.42}\\ 
horse & 29.4 & 35.10 & 41.84 & 44.82 & \textbf{54.52}\\ 
mbike & 21.3 & \textbf{31.61} & 26.70 & 27.07 & 29.77\\ 
train & 13.2 & \textbf{39.58} & 20.48 & 24.93 & 29.23\\ 
\hline
mAP & 20.57 & 29.41 & 29.67 & 31.92 & \textbf{37.413}\\ 
\hline
\end{tabular}
\setlength{\abovecaptionskip}{15pt plus 3pt minus 2pt} 
\caption{Object Detection Results on Youtube-Objects test set. Pre-trained R-CNN detector\cite{girshick14CVPR} is downloaded from \cite{R-CNN-model}. Detection result using Deformable Parts Model \cite{DPM10} (DPM) are from \cite{KalogeitonFS15}. Fine-tune SS uses fine-tuning with Selective Search proposals (similar to R-CNN), Fine-tune EB uses fine-tuning with EdgeBoxes proposals and Fine-tune VOP uses fine-tuning with proposed video Object Proposals ($\lambda = 0.2$).} 
\label{table:youtube-detection}
\squeezeup 
\end{table}

Fine-tuning on Youtube-Objects training data with Selective Search proposals \cite{SelectiveSearch2013IJCV} improves the detection results by at least $9$\%\ compared with the model fine-tuned for the image dataset PASCAL VOC. The detector learned with EdgeBoxes performs better than the one learned with Selective Search proposals. However, the detector learned with VOP even outperforms the detection rate by another $5.5$\%\ and achieves state-of-the-art detection accuracy ($37.4$\%) on this dataset. Although detection accuracy for ``cat'', ``cow'', ``dog'' and ``horse'' have improved by a huge margin after using CNN features, categories like ``train'' and ``bird'' are still best detected using DPM \cite{DPM10} detector. 

\begin{table*}[t]
\centering
\begin{tabular}{|c| c|c| c|c| c|c|c|c| c|c|c|c|}
\hline
\multirow{3}{*}{} & \multirow{3}{*}{PF SS} & \multirow{3}{*}{PF EB} & \multicolumn{2}{|c|}{PF VOP} &\multicolumn{8}{|c|}{OVERLAP}\\ \cline{4-13}
& & & \multirow{2}{*}{CPU OF} & \multirow{2}{*}{GPU OF} & \multicolumn{4}{|c|}{CPU Optical Flow} & \multicolumn{4}{|c|}{GPU Optical Flow}\\ \cline{6-13}
& & & & &200 V &500 V&1K V &2K V &200 V &500 V &1K V &2K V\\ \hline
Prop time & 10 & 0.3 & 3.8 & 1.3 & \multicolumn{4}{|c|}{3.8} & \multicolumn{4}{|c|}{1.3}\\ \hline
Overall time & 30 & 20.3 & 23.8 & 21.3 & 6.2 & 9.3& 14.6 & 28.0 & 3.7 & 6.8 &12.1 & 26.5\\ \hline
mAP & 29.62 & 31.95 & 37.72 & 37.72 & 28.59 & 33.59 & 35.82 & 36.63 & 28.59 & 33.59 & 35.82 & 36.63\\ \hline 
\end{tabular}
\setlength{\abovecaptionskip}{15pt plus 3pt minus 2pt} 
\caption{Complexity and accuracy comparison. Proposal generation (Prop time),  overall detection time for per-frame (PF) baseline methods with Selective Search (SS) proposals, EdgeBoxes (EB) proposals, proposed VOPs and OVERLAP are shown. Baseline methods use 2000 proposals per frame. GPU OF and CPU OF denote GPU-based and CPU-based optical flow (OF) respectively. mAP increases as number of VOPs increases. About 3$\times$ speedup achived with 500 VOPs with only 4\%\ drop in mAP compared to the baseline per-frame detection.} 
\label{table:OVERLAP-comparison}
\end{table*}




\subsection{Streaming Clustering of VOP }
\begin{figure}
\begin{center}
	\includegraphics[scale = 0.5]{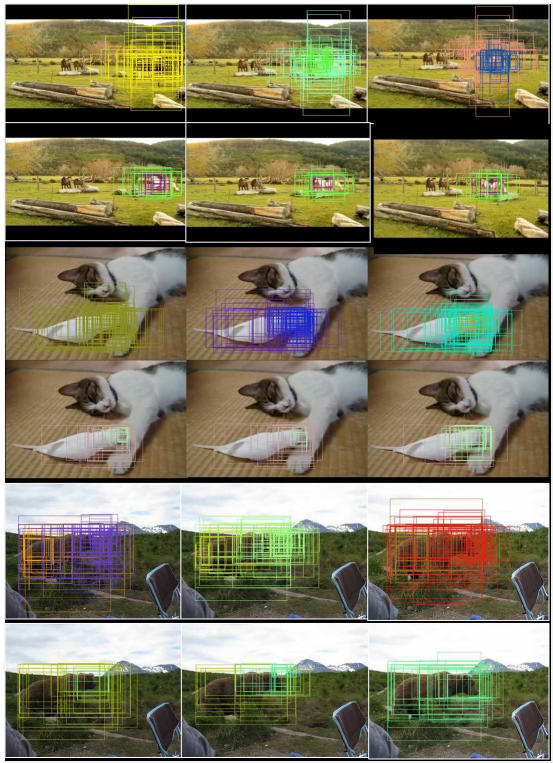}
\end{center}
   \caption{Temporal consistency in streaming clustering of VOPs on arbitrary videos ``Horse riding'', ``Bird-cat'' and ``Alaskan bear'' downloaded from the Internet. Windows drawn in same color belong to same cluster. Top rows of every pair show the proposals clustered on individual frame (frames \#2, \#10, \#25 in each case) level; bottom row shows the results of streaming clustering. }
\label{fig:Streaming-VOP}
\squeezeup 
\end{figure}

Figure \ref{fig:Streaming-VOP} shows the results of frame-level clustering vs streaming clustering at sub-sequence levels on arbitrary videos downloaded from Youtube. In these experiments, a sub-sequence contains 3 video frames, with an overlap of one frame from the previous sub-sequence. We use high $\lambda$ value ( $0.8$ ) to identify only the moving objects with very few number of proposals. For clear visualization, we use only 50 proposal windows at every frame and that makes less than 200 proposals per sub-sequence. We aim to take the advantage of fast approximate spectral clustering algorithm which scales linearly with the problem size. In our current implementation, clustering takes less than 0.1 second for 3 frames streaming-volume with 50 VOPs per frame.


\subsection{Video Object Detection}
To investigate the relative detection rate with OVERLAP compared with frame-wise R-CNN like approach, we create a subset (955 frames) of Youtube-Objects test-set (1783 frames) where the video frames form a valid video play. We find that for OVERLAP, CNN feature extraction and classification is needed only for $10$\%\ to $30$\%\ windows among all of them. Table \ref{table:OVERLAP-comparison} corroborates the fact that the detection accuracy improves as we increase the number of VOPs from 200 to 2000 in OVERLAP at the cost of increased complexity needed for spectral clustering. Difference in mAP is between $1$-$9$\%\ . As an example, compared to per-frame detection, OVERLAP achieves about 3$\times$ speedup at the cost of only 4\% mean Average Precision (mAP) with 500 VOPs per frame. The non-GPU based spectral clustering implementation in MATLAB makes cases for more than 2000 VOPs even slower than per-frame RCNN. 

Per-frame proposal generation with Selective Search and EdgeBoxes takes about 10 seconds \cite{Hosang2015pami} and 0.3 seconds \cite{Hosang2015pami} respectively. Overall time for object detection per-frame in R-CNN per-frame becomes about 30 seconds and 20.3 seconds with the above corresponding methods. Generation of VOPs requires optical flow which takes 3.5 sec per frame in CPU-implementation and 1 sec per frame for GPU-implementation for about $500\times500$ resolution frame-pairs. ``CPU'' and ``GPU'' in Table \ref{table:OVERLAP-comparison} denote optical flow implementation in CPU \cite{TBroxOF11} and GPU \cite{GPU_Optical_flow_10} respectively. 
The Youtube-Objects dataset  mostly contains moving objects. In addition, the test dataset does not contain significant number of test cases where multiple instances of objects with similar appearances are spatially overlapped. Thus, accuracy of OVERLAP successfully approaches the baseline per-frame detection accuracy as we increase the number of VOPs. For the fastest detection (with 200 VOPs only), we use more weight (lambda = 0.6) for temporal edge and still manage to get acceptable detection accuracy with over 5$\times$ speedup as shown in the column corresponding to 200 VOP in Table \ref{table:OVERLAP-comparison}. 
GPU-based spectral clustering can potentially lead to further speed up.



\squeezeup 

\section{Conclusion}
Experimental results show that VOP helps in learning a better moving or static video object detector model and achieves state-of-the-art detection accuracy on Youtube-Video dataset. We show that the proposed OVERLAP framework can detect temporally consistent objects in videos through object class label propagation using streaming clustering of VOPs with significant  speedup compared with naive per-frame detection with acceptable loss of accuracy. We also show that multiple objects segmentation can also be achieved as a by-product of OVERLAP.

{\small
\bibliographystyle{ieee}
\bibliography{egpaper_final}
}

\end{document}